\documentclass{article}

\usepackage[preprint]{neurips_2026}

\usepackage[utf8]{inputenc}
\usepackage[T1]{fontenc}
\usepackage{hyperref}
\usepackage{url}
\usepackage{booktabs}
\usepackage{amsfonts}
\usepackage{nicefrac}
\usepackage{microtype}
\usepackage{graphicx}
\usepackage{xcolor}

\title{What Should an Agent Remember? Disentangling Retention from\\ Retrieval in Bounded-Memory Evaluation}

\author{%
  Juli Huang \\
  Stanford University \\
  \texttt{julih@stanford.edu} \\
  \thanks{Code: \url{https://github.com/TheClassicTechno/MemoryLLMAgentProject}}
}

\begin{document}

\maketitle

\begin{abstract}
A persistent agent must decide both \emph{what to retain} as information arrives and \emph{what to
surface} once a query appears, different decisions made at different times with different
information, yet memory evaluations can end up comparing methods that differ on both at once. We build a
streaming-recall environment crossing a retention rule with a selection rule, scoring the same 300
seeded episodes in every cell. Holding access fixed, query-aware selection buys $+15.5$ percentage
points (pp) of required-fact recall (95\% CI $[+12.8, +18.2]$), whereas the mixed-cell contrast used
by an earlier version of this evaluation reports $+68.7$ pp $[+65.2, +72.0]$; the extra $+53.2$ pp is access.
With query-aware, dense, or oracle selection, recall in every bounded retention condition then
exactly equals its retention rate: a 768-dimensional dense retriever moves it by $+0.00$ pp $[+0.00, +0.00]$, in the bounded recency-retention condition all
319 observed retrieval failures are evictions rather than ranking errors, and recall there falls to
$0\%$ as the target recedes into the past. In that synthetic evaluation, incidental benchmark properties can create substantial spurious
advantages; rerunning the identical grid on SQuAD reproduces the ceiling in all 25 cells, though
there the dense retriever does beat the lexical one, so that particular null was a property of our
generator. Bounded-memory
evaluations should hold access fixed and report retention and selection separately.
\end{abstract}

\section{Introduction}

Long-running agents can accumulate more observations than they can keep available at once, so some
must be dropped. This makes memory a control problem: given a budget of $K$ items, which
observations remain available later? Existing memory architectures each embed an assumption about
what is worth keeping: that recent, repeated, or query-resembling things matter.

Evaluating those assumptions is harder than it looks, because an agent makes two decisions. A
\textbf{retention} decision happens online, as each observation arrives, and is necessarily
\emph{query-independent}: the future query has not been seen. A \textbf{selection} decision happens at
query time and \emph{can} be query-aware. A method that appears to win may be winning at either stage,
or may simply have been granted access to material its competitors discarded. One accuracy number per
method cannot distinguish these.

This is easy to introduce accidentally. Auditing an earlier version of this evaluation, which
compared five memory heuristics, we found that its strongest methods re-ranked the complete history at query time while its weakest saw
only a bounded buffer; that a template defect left its queries with no token identifying the fact
asked about; and that its distractors carried a literal marker any overlap scorer could exploit. Two
of its five conditions in fact retained the entire history and so shared an access regime, yet were
reported far apart, a gap that cannot reflect memory-budget behaviour at all. Each of these is mundane
on its own. Together they produced a large, clean, misleading result.

We therefore rebuild the evaluation as a factorial design. Our contributions:
\begin{itemize}\itemsep2pt \parskip0pt \parsep0pt
\item A streaming-recall benchmark that crosses \textbf{retention rule} $\times$ \textbf{selection
rule}, evaluates every cell on the same seeded episodes, and reports recall alongside the retention
rate that upper-bounds it (\S\ref{sec:method}).
\item A decomposition showing that most of the apparent advantage of query-aware memory in the
mixed-cell comparison is attributable to history access rather than query awareness, with the
remainder capped by retention (\S\ref{sec:decomp}).
\item Evidence that query-aware retrieval gives no protection against delay-induced eviction; that
the retention ceiling is selection-independent, verified against a dense retriever rather than
assumed; and that incidental benchmark properties can create spurious apparent gains, with the
stronger retriever amplifying one such artifact (\S\ref{sec:pressure}, \S\ref{sec:artifacts}); and that the ceiling result
reproduces on real SQuAD text while the dense-versus-lexical null does not (\S\ref{sec:squad}).
\end{itemize}

\section{Related work}

Agent memory systems score stored items by a mixture of signals: \citet{park2023generative} retrieve
from a memory stream by weighted recency, importance, and relevance; \citet{packer2023memgpt} page
between a bounded context and external storage; \citet{zhong2024memorybank} add a forgetting
schedule. These systems couple several memory-management decisions within end-to-end evaluations,
which is exactly the regime in which retention and retrieval effects become hard to separate.

The selection half is classical retrieval, where lexical scorers such as BM25
\citep{robertson2009probabilistic} remain strong against dense retrievers
\citep{karpukhin2020dense} and RAG \citep{lewis2020retrieval} conditions generation on retrieved
evidence. But that literature assumes a fixed corpus, with nothing thrown away; bounded memory breaks
that assumption. Relatedly, \citet{liu2024lost} show that where evidence sits in a context changes
whether a model uses it: another case in which an apparent capability difference turns out to be a
property of what was made available to the model. We contribute to the evaluation layer rather than
proposing a memory architecture; our object of study is what a bounded-memory benchmark measures.

\section{A factorial streaming-recall benchmark}
\label{sec:method}

\paragraph{Environment.} An episode is a stream of 10 facts followed by 2 queries. Every fact has the
identical surface form \texttt{"The \textit{slot} is \textit{value}."}; slots are drawn without
replacement within an episode, so a query naming a slot identifies exactly one fact. A fraction
$d$ (default $0.3$) of facts are distractors, drawn from a disjoint slot vocabulary but written in
the same form, so distractors are \emph{not} lexically marked. Each query targets a uniformly
sampled non-distractor fact.

\paragraph{The two factors.} A \textbf{retention rule} consumes the stream online under capacity $K$
and is never shown the query: \textsc{recency} (keep the last $K$), \textsc{random} (admit with
$p{=}0.5$, FIFO evict), \textsc{frequency} (evict the item whose tokens are rarest so far, FIFO on
ties), \textsc{type-oracle} (admit only non-distractors; this reads the generator's ground-truth
label and is therefore \emph{privileged}, not implementable by an agent), and \textsc{unbounded}
(keep everything; the no-forgetting reference). A \textbf{selection rule} then ranks the survivors
and exposes the top $k{=}3$: \textsc{recency}, \textsc{random}, \textsc{query} (Jaccard overlap
between query and fact tokens, recency tie-break), \textsc{dense} (cosine similarity between MPNet
sentence embeddings\footnote{\texttt{all-mpnet-base-v2}, 768-d. The vocabularies are closed, so we cache all
353 realizable strings once; selection is then a dot product, and exactly deterministic.}), and \textsc{oracle}
(surface the target if present). We call the Jaccard rule \emph{query-similarity} and never an
embedding method; \textsc{dense} is the only rule here that uses learned vectors.

\paragraph{Query informativeness.} We also vary whether the query carries usable signal. An
\emph{informative} query is \texttt{"What is the \textit{slot}?"} naming the target's slot; an
\emph{uninformative} query is \texttt{"What is the value?"}, where \texttt{value} appears in no fact.
The two differ by exactly one token, isolating query informativeness from all else.

\paragraph{Metric and statistics.} We report \textbf{required-fact recall}: the fraction of queries
whose target fact appears among the $k$ exposed items. We report recall rather than answer accuracy
because it isolates the memory pipeline; \S\ref{sec:llm} then puts a language model at the end
and shows recall predicts answer correctness up to a near-constant factor. Alongside recall we report the
\textbf{retention rate}, the fraction of queries whose target survived retention, which
upper-bounds recall for any selection rule. All cells use the same 300 seeded episodes (600
queries); confidence intervals are percentile bootstrap over 2{,}000 resamples \emph{of episodes},
since two queries from one stream are not independent. Seed, configuration, and per-query records
are saved with the results.

\section{Results}

\subsection{RQ1: how much of the advantage is query awareness?}
\label{sec:decomp}

Table~\ref{tab:matrix} gives the full grid under informative queries. Reading it by column isolates
selection; by row, retention.

\begin{table}[t]
\centering
\caption{Required-fact recall (\%) under \emph{informative} queries, $K{=}5$, $k{=}3$, 600 queries per cell on the same 300 episodes. ``Retention'' is the fraction of queries whose target survived retention and upper-bounds every entry in its row. $^{\dagger}$\textsc{type-oracle} reads the generator's labels and is privileged, not implementable by an agent. The 100\% entries are a construction ceiling, not a performance result: an informative query token-matches exactly one fact, so selection is exactly solvable when nothing was discarded.}
\label{tab:matrix}
\small
\begin{tabular}{lrrrrrr}
\toprule
Retention rule & \textsc{random} & \textsc{recency} & \textsc{query} & \textsc{dense} & \textsc{oracle} & Retention \\
\midrule
\textsc{unbounded} & 31.5 & 31.3 & 100.0 & 100.0 & 100.0 & 100.0 \\
\textsc{recency} & 27.0 & 31.3 & 46.8 & 46.8 & 46.8 & 46.8 \\
\textsc{random} & 27.8 & 30.8 & 46.8 & 46.8 & 46.8 & 46.8 \\
\textsc{frequency} & 26.8 & 30.2 & 45.8 & 45.8 & 45.8 & 45.8 \\
\textsc{type-oracle}$^{\dagger}$ & 43.7 & 41.2 & 70.3 & 70.3 & 70.3 & 70.3 \\
\bottomrule
\end{tabular}
\end{table}

Two things follow. First, in every bounded row the \textsc{query}, \textsc{dense} and
\textsc{oracle} columns are \emph{exactly equal} to the retention rate ($46.8$, $46.8$, $45.8$,
$70.3$): selection saturates \emph{in this environment}, and every remaining failure is a retention
failure. We test that with a far stronger scorer rather than asserting it: a 768-dimensional MPNet
retriever changes recall by $+0.00$ pp $[+0.00, +0.00]$ against both the lexical rule and an oracle
selector, under bounded and unbounded retention alike, and across all 50 cells no selection rule
exceeded its row's retention rate. Saturation is a property of this generator, not a general one:
\S\ref{sec:squad} shows that on real text a dense retriever does improve selection, and yet still
cannot pass the retention ceiling. The ceiling is the general claim; saturation is not. A better retriever is not merely
unhelpful; it is arithmetically incapable of helping, the exposed set being a subset of what
retention kept.

\begin{figure}[t]
\centering
\includegraphics[width=\textwidth]{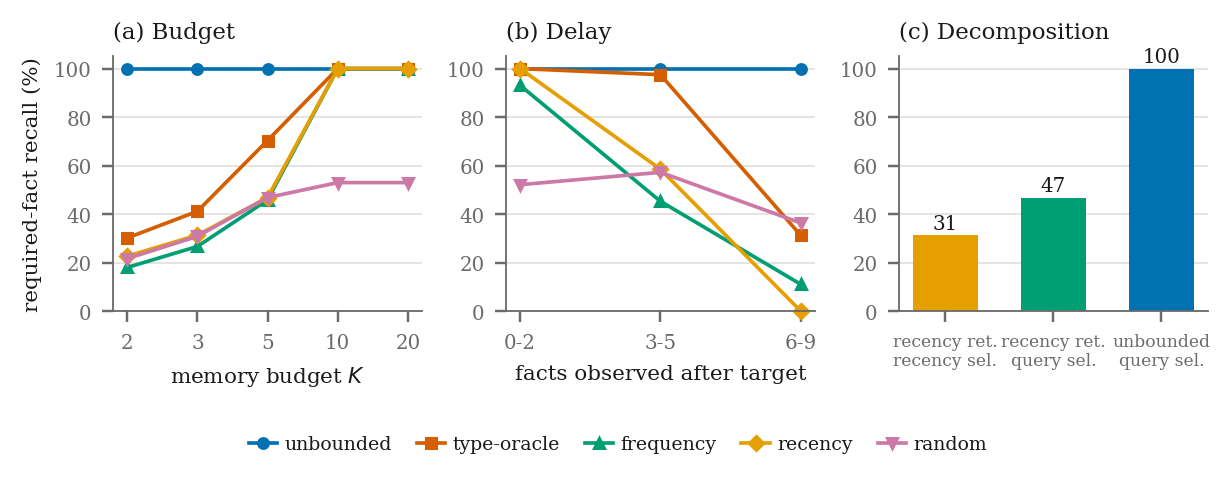}
\caption{Required-fact recall under query-aware selection; the lexical and dense scorers are
identical throughout under informative queries, so one set of curves serves for both.
\textbf{(a)} vs.\ memory budget $K$;
selection is inert at $K \le k{=}3$. \textbf{(b)} vs.\ delay (facts observed after the target;
$n{=}188/159/253$): recency retention falls to \emph{zero}: query-aware retrieval offers no
protection against eviction. \textbf{(c)} The apparent advantage at $K{=}5$: adding query-aware selection at
matched access gives $+15.5$ pp, adding full-history access a further $+53.2$ pp.}
\label{fig:main}
\end{figure}

Second, a mixed-cell comparison does not report the contrast it appears to.
\textsc{unbounded}+\textsc{query} against \textsc{recency}+\textsc{recency}, structurally what our
earlier evaluation did, gives $+68.7$ pp $[+65.2, +72.0]$; fixing access at a bounded buffer and
varying only selection gives $+15.5$ pp $[+12.8, +18.2]$; fixing selection and varying only access
gives $+53.2$ pp $[+49.7, +56.8]$. Three-quarters of the headline effect is the freedom to re-rank
material the baselines had discarded (Fig.~\ref{fig:main}c).

\subsection{RQ2: behaviour under memory pressure}
\label{sec:pressure}

\begin{table}[t]
\centering
\caption{Recall (\%) under query-aware selection across memory budgets $K$, informative queries,
$k{=}3$. At $K \le k$ nothing retained is withheld, so selection is inert and every rule ties.}
\label{tab:budget}
\small
\begin{tabular}{lrrrrr}
\toprule
Retention rule & $K{=}2$ & $K{=}3$ & $K{=}5$ & $K{=}10$ & $K{=}20$ \\
\midrule
\textsc{unbounded} & 100.0 & 100.0 & 100.0 & 100.0 & 100.0 \\
\textsc{recency} & 22.7 & 31.3 & 46.8 & 100.0 & 100.0 \\
\textsc{random} & 21.7 & 30.8 & 46.8 & 53.0 & 53.0 \\
\textsc{frequency} & 18.0 & 26.7 & 45.8 & 100.0 & 100.0 \\
\textsc{type-oracle}$^{\dagger}$ & 30.2 & 41.2 & 70.3 & 100.0 & 100.0 \\
\bottomrule
\end{tabular}
\end{table}

\paragraph{Budget.} Recall under recency retention with query-aware selection rises
$22.7 \to 31.3 \to 46.8 \to 100 \to 100$ across $K \in \{2,3,5,10,20\}$ (Table~\ref{tab:budget},
Fig.~\ref{fig:main}a). Below $K \le k$ the selection rule is inert, because everything retained is
already exposed, so the entire spread across selection rules at small budgets is zero. This deserves
stating plainly: \emph{a query-aware retriever can only matter when the buffer is larger than the
retrieval budget.} Every bounded rule converges to the unbounded ceiling once $K$ reaches the stream
length, which is the trivial regime in which forgetting has stopped happening at all.

\paragraph{Delay.} Bucketing queries by how many facts arrived after the target
(Fig.~\ref{fig:main}b), recency retention yields $100\% \to 58.5\% \to 0.0\%$ across delays of
$0$--$2$, $3$--$5$, and $6$--$9$ facts, and type-oracle retention $100\% \to 97.5\% \to 31.2\%$;
unbounded retention is flat at $100\%$ (Table~\ref{tab:delay}). The collapse to exactly zero is the
sharpest result we have. Selection is oracle-equivalent in every one of these rows, so this is not a
retrieval failure at all: by the time the query arrives the fact is gone, and no scorer, however
strong, can rank an item that is no longer there. Filtering distractors (type-oracle) delays the
cliff but does not remove it, falling to $31.2\%$ at the longest delay.

\begin{table}[t]
\centering
\caption{Recall (\%) under query-aware selection by delay, measured as the number of facts observed
after the target, $K{=}5$, informative queries ($n = 188 / 159 / 253$ queries per bucket). Selection is
oracle-equivalent in every row, so the decline is retention loss, not retrieval failure.}
\label{tab:delay}
\small
\begin{tabular}{lrrr}
\toprule
Retention rule & 0--2 & 3--5 & 6--9 \\
\midrule
\textsc{unbounded} & 100.0 & 100.0 & 100.0 \\
\textsc{recency} & 100.0 & 58.5 & 0.0 \\
\textsc{random} & 52.1 & 57.2 & 36.4 \\
\textsc{frequency} & 93.1 & 45.3 & 11.1 \\
\textsc{type-oracle}$^{\dagger}$ & 100.0 & 97.5 & 31.2 \\
\bottomrule
\end{tabular}
\end{table}

\begin{table}[t]
\centering
\caption{Recall (\%) under query-aware selection as distractor density $d$ rises, $K{=}5$,
informative queries. Higher density \emph{benefits} the privileged type-oracle, since more of the
stream is filtered before capacity is consumed, while recency retention is roughly unchanged.}
\label{tab:density}
\small
\begin{tabular}{lrrrr}
\toprule
Retention rule & $d{=}0.1$ & $d{=}0.3$ & $d{=}0.5$ & $d{=}0.7$ \\
\midrule
\textsc{unbounded} & 100.0 & 100.0 & 100.0 & 100.0 \\
\textsc{recency} & 47.3 & 46.8 & 46.5 & 50.5 \\
\textsc{random} & 44.8 & 46.8 & 41.0 & 44.2 \\
\textsc{frequency} & 47.5 & 45.8 & 35.3 & 23.5 \\
\textsc{type-oracle}$^{\dagger}$ & 54.0 & 70.3 & 100.0 & 100.0 \\
\bottomrule
\end{tabular}
\end{table}

\paragraph{Distractor density.} Raising density from $0.1$ to $0.7$ leaves recency retention flat
($47.3 \to 50.5$) but lifts type-oracle retention from $54.0$ to $100.0$ (Table~\ref{tab:density}),
because discarding a larger share of the stream leaves proportionally more room for targets.
Raising density therefore \emph{benefits} the privileged type-oracle, because more of the stream can
be filtered before capacity is consumed, while recency retention is left roughly unchanged. The rule
that gains is the one that needs the answer key to do it.

\subsection{RQ3: benchmark properties that create spurious apparent gains}
\label{sec:artifacts}

\begin{table}[t]
\centering
\caption{The same grid under \emph{uninformative} queries, which name no fact. The \textsc{query} column collapses onto \textsc{recency} exactly: with no discriminative token every fact ties and the rule degenerates to its tie-break. \textsc{dense} does not collapse, but gains nothing from the query either (\S\ref{sec:artifacts}).}
\label{tab:uninformative}
\small
\begin{tabular}{lrrrrrr}
\toprule
Retention rule & \textsc{random} & \textsc{recency} & \textsc{query} & \textsc{dense} & \textsc{oracle} & Retention \\
\midrule
\textsc{unbounded} & 31.5 & 31.3 & 31.3 & 42.0 & 100.0 & 100.0 \\
\textsc{recency} & 27.0 & 31.3 & 31.3 & 35.8 & 46.8 & 46.8 \\
\textsc{random} & 27.8 & 30.8 & 30.8 & 39.3 & 46.8 & 46.8 \\
\textsc{frequency} & 26.8 & 30.2 & 30.2 & 35.2 & 45.8 & 45.8 \\
\textsc{type-oracle}$^{\dagger}$ & 43.7 & 41.2 & 41.2 & 39.7 & 70.3 & 70.3 \\
\bottomrule
\end{tabular}
\end{table}

\begin{table}[t]
\centering
\caption{Distractor marking control, \emph{uninformative} queries only. Prefixing distractors with a
literal marker string, as the original generator did, lifts the query-aware rule by $+9.8$ pp $[+7.7, +12.2]$ and $+9.0$ pp $[+7.0, +11.3]$ while
leaving every other rule unchanged. The scorer is detecting the marker, not matching the query.}
\label{tab:marking}
\small
\begin{tabular}{llrrr}
\toprule
Retention & Distractors & \textsc{recency} & \textsc{query} & \textsc{oracle} \\
\midrule
\textsc{unbounded} & unmarked & 31.3 & 31.3 & 100.0 \\
\textsc{unbounded} & marked & 31.3 & 41.2 & 100.0 \\
\textsc{recency} & unmarked & 31.3 & 31.3 & 46.8 \\
\textsc{recency} & marked & 31.3 & 40.3 & 46.8 \\
\bottomrule
\end{tabular}
\end{table}

Under \emph{uninformative} queries (Table~\ref{tab:uninformative}) the \textsc{query} column
collapses onto the \textsc{recency} column exactly ($31.3$, $31.3$, $30.8$, $30.2$, $41.2$ down the rows): with no discriminative token,
all facts tie and the rule degenerates to its tie-break. Query-aware selection contributes literally
nothing, which is the regime our earlier evaluation was unknowingly in.

Yet an apparent advantage can be restored without restoring any query signal, in two ways. Re-marking
distractors with a literal prefix, as the original generator did, raises uninformative-query recall
for the query-aware rule by $+9.8$ pp $[+7.7, +12.2]$ unbounded and $+9.0$ pp $[+7.0, +11.3]$ bounded,
leaving every other rule unchanged (Table~\ref{tab:marking}): the scorer is detecting the marker,
not matching the query.

The dense retriever reproduces this with no marker at all. Under uninformative queries it does
\emph{not} collapse onto \textsc{recency}, reaching $42.0\%$ against a $31.5\%$ chance level, readable
as embeddings extracting signal where lexical overlap cannot. They do not: $0.0\%$ of the items it
exposes are distractors, against a $30\%$ base rate. It filters distractors from the benchmark's latent type
structure, then picks at random among the $7$ relevant facts: $3/7=42.9\%$ predicted, $42.0\%$
observed. The stronger retriever amplifies the artifact rather than resisting it.

\subsection{RQ4: does recall predict answer correctness?}
\label{sec:llm}

Recall is a retrieval-side proxy. To test whether it tracks task success we place FLAN-T5-large at the
end of the pipeline (greedy decoding, normalized exact match, 300 queries from the same episode set) and score its
answers (Table~\ref{tab:llm}).

\begin{table}[t]
\centering
\caption{Answer accuracy (\%) with FLAN-T5-large at the end of the pipeline, greedy decoding,
300 queries per condition drawn from the same seeded episodes, normalized exact
match. ``Acc.\ $\mid$ present'' conditions on the required fact having reached the model and is
close to constant across conditions, which is what licenses recall as the primary metric.}
\label{tab:llm}
\small
\begin{tabular}{lrrr}
\toprule
Context given to the model & Recall & Answer acc. & Acc.\ $\mid$ present \\
\midrule
No memory & 0.0 & 0.0 & --- \\
Bounded ($K{=}5$), \textsc{recency} selection & 30.7 & 23.3 & 70.7 \\
Bounded ($K{=}5$), \textsc{query} selection & 47.3 & 36.3 & 73.2 \\
Full context (all 10 facts) & 100.0 & 73.7 & 73.7 \\
Required fact alone & 100.0 & 74.7 & 74.7 \\
\bottomrule
\end{tabular}
\end{table}

 With no memory the model scores $0.0\%$, so the task genuinely requires memory. A bounded
agent ($K{=}5$, query selection) recalls the target for $47.3\%$ of queries and answers $36.3\%$
correctly; full context scores $73.7\%$ and the required fact alone $74.7\%$, so nine irrelevant facts
cost $1.0$ pp and the bottleneck is memory, not context noise. Accuracy conditioned on the fact having
reached the model spans only $70.7$--$74.7\%$, so under these tested conditions answer accuracy is
recall scaled by a roughly constant factor, which is what licenses recall as the primary metric. The bounded agent's $63.7\%$ failure rate
splits into $52.7$ pp where the fact never arrived and $12.7$ pp where it arrived and the model erred
anyway: five in six failures are memory failures.

\subsection{RQ5: what explains the remaining failures?}
\label{sec:errors}

We classify every failure of the bounded agent (recency retention, query selection) into
memory-specific categories: eviction, where the target never survived retention; displacement, where
it was retained but pushed out of the top-$k$; and similarity confusion, where a distractor outscores
the target under the query scorer itself. All 319 failures are evictions. Displacement and similarity
confusion account for zero. Swapping in a query-independent selector on the same streams adds 93
displacement failures, so query awareness removes the selection-stage failure mode outright while
leaving the dominant one untouched. With the language model attached, of 300 queries the target was
absent 158 times and present-but-answered-wrongly 38 times; the latter is the only failure here that
is not about memory, and it matches the $12.7$ pp answering term of \S\ref{sec:llm}.

\section{Does this survive real text?}
\label{sec:squad}

Everything so far runs on a generator we wrote, which is the obvious objection. We therefore rerun
the identical grid on SQuAD v1.1 dev \citep{rajpurkar2016squad}, changing only where episodes come
from. For each of 300 questions we split the target paragraph into sentences, mark the sentence
containing the answer span as the target, and pad the stream to ten sentences with sentences drawn
from unrelated paragraphs. The distractors are ordinary English on other topics, so nothing marks
them, and the query is the real SQuAD question rather than a template. The retention rules, selection
rules, metric and bootstrap are imported and called unmodified.

\begin{table}[t]
\centering
\caption{The same grid on \emph{real SQuAD text}: recall (\%) of the answer-bearing sentence, 300 questions from SQuAD v1.1 dev, $K{=}5$, $k{=}3$. Each stream mixes the target paragraph's sentences with sentences from unrelated paragraphs, so nothing marks the distractors. The retention ceiling holds in all 25 cells. Unlike the synthetic benchmark \textsc{dense} now beats \textsc{query}, yet under a bounded buffer it still only reaches the ceiling, never past it.}
\label{tab:squad}
\small
\begin{tabular}{lrrrrrr}
\toprule
Retention rule & \textsc{random} & \textsc{recency} & \textsc{query} & \textsc{dense} & \textsc{oracle} & Retention \\
\midrule
\textsc{unbounded} & 31.7 & 31.3 & 91.7 & 97.7 & 100.0 & 100.0 \\
\textsc{recency} & 29.3 & 31.3 & 51.7 & 53.3 & 53.3 & 53.3 \\
\textsc{random} & 25.3 & 29.3 & 39.3 & 39.7 & 40.0 & 40.0 \\
\textsc{frequency} & 40.7 & 42.3 & 63.3 & 64.3 & 65.3 & 65.3 \\
\textsc{type-oracle}$^{\dagger}$ & 64.0 & 69.3 & 86.3 & 87.7 & 89.3 & 89.3 \\
\bottomrule
\end{tabular}
\end{table}

The central claim survives, and one secondary claim does not. The retention ceiling holds in all 25
cells: no selection rule anywhere exceeds its row's retention rate (Table~\ref{tab:squad}). Under a
bounded recency buffer the dense retriever again lands exactly on the ceiling, $+0.00$ pp
$[+0.00, +0.00]$ against an oracle selector, and unbounded access is worth $+44.3$ pp
$[+38.3, +50.3]$ to it. Query awareness at matched access is worth $+20.3$ pp $[+15.7, +25.0]$, more
than the $+15.5$ pp we measured synthetically, which is what one would expect once queries stop being
templates.

What does \emph{not} transfer is the dense-versus-lexical null. On real text the dense retriever
genuinely beats Jaccard, by $+6.0$ pp $[+3.0, +9.3]$ with unbounded retention and $+1.7$ pp
$[+0.3, +3.3]$ under a bounded buffer. Our synthetic finding that the two are exactly equivalent was
therefore a property of that generator, where an informative query token-matches its target exactly,
and not a general fact. The dense-versus-lexical equivalence observed synthetically therefore does
not generalize to natural text: a better retriever does help at the selection stage there. It still cannot exceed what retention
kept, which is the claim the paper actually makes.

\section{Limitations}

The main environment is synthetic, single-episode and small (10 facts, 2 queries), built so selection
is exactly solvable under informative queries, hence the 100\% ceilings; \S\ref{sec:squad} rebuts the
strongest form of that objection but both environments are still single-turn recall rather than
multi-session agent memory. That isolates retention, but
means our numbers \emph{bound} what selection contributes rather than estimating it on a realistic
corpus; with paraphrase, partial overlap, or multi-hop queries the matched-access query-awareness
effect would likely be larger than the $+15.5$ pp we measure. The main grid is retrieval-side, and
the answer-accuracy check of \S\ref{sec:llm} uses a single small model with one prompt and greedy
decoding, so it establishes that recall tracks answer correctness in this setting rather than
characterizing how any particular agent would behave. Our ``delay'' is
positional distance within a stream, not elapsed time or cross-session persistence: this is bounded
streaming memory, not long-term memory. A cross-encoder reranker could differ from our two scorers,
though the ceiling argument applies to any selector. We excluded rather than reported an untrained
linear write-gate in the repository: it has no held-out split and its reward comes from the answer key.

\section{Conclusion}

Retention and selection are different decisions, and one accuracy number cannot tell them apart. Hold
access fixed across compared methods, and report the retention rate beside every recall number. When the required
memory has been evicted a better retriever cannot recover it: retrieval performance is upper-bounded
by retention.

\bibliographystyle{plainnat}

\end{document}